\documentclass[runningheads]{llncs}
\usepackage[T1]{fontenc}
\usepackage{graphicx,verbatim}
\usepackage{kotex}
\usepackage{booktabs}
\usepackage{multirow}
\usepackage{amssymb}
\usepackage{amsmath}
\usepackage[table]{xcolor}
\usepackage{lipsum}

\usepackage{colortbl}
\usepackage{xcolor}
\usepackage{marvosym}

\begin{document}
%

\title{Leveraging Pathology Co-occurrence for Test-Time Adaptation in Chest X-Ray Diagnosis}

\titlerunning{Pathology Co-occurrence for Test-Time Adaptation}
%

\author{Woojin Jeong\inst{1}$^{\dagger}$ \and
Yujin Choi\inst{2, 3}$^{\dagger}$ \and
Dongbin Kim\inst{1} \and
Soyeon Park\inst{1} \and
Jaewook Lee\inst{1}\textsuperscript{(\Letter)}}
%
\authorrunning{W. Jeong et al.}
\institute{Seoul National University, 1, Gwanak-ro, Seoul, 08826, Republic of Korea \and
Nanyang Technological University, 50 Nanyang Avenue, 639798, Singapore \and UNIST, 50, UNIST-gil, Ulsan, 44919, Republic of Korea \\
\email{jaewook@snu.ac.kr}}

\maketitle              
\let\thefootnote\relax
\footnotetext[0]{$^{\dagger}$ Equal contribution.}

%

\begin{abstract}
Medical imaging models often degrade when deployed at new clinical sites due to differences in imaging equipment, protocols, and patient populations. Test-time adaptation (TTA) addresses this by updating a pretrained model using only unlabeled target data, without access to source data. However, existing TTA methods were designed for single-label classification on natural image benchmarks, minimizing entropy uniformly across all samples without considering label dependencies. This overlooks a key property of multi-label medical imaging: pathologies do not occur independently but exhibit structured co-occurrence patterns. In this work, we propose \textbf{Co}-occurrence \textbf{W}eighted \textbf{A}daptation (CoWA), which leverages disease co-occurrence patterns as a reliability signal for adaptation. CoWA estimates label co-occurrence structure from model predictions and downweights samples that deviate from expected patterns, enabling adaptation to rely more on consistent predictions while reducing the impact of noisy ones. We evaluate CoWA on chest X-ray benchmarks under domain shifts and demonstrate consistent improvements over established baselines. The code is available at \url{https://github.com/woojin716/CoWA}.
\end{abstract}

\keywords{Chest X-ray \and Domain Shift \and Test-time Adaptation}

%
%
%

\section{Introduction}
Deep learning models for chest X-ray classification have demonstrated strong performance on benchmark datasets~\cite{irvin2019chexpert,wang2017chestx}. In clinical practice, however, models are often deployed across institutions where imaging equipment, acquisition protocols, and patient demographics differ from those in the training environment~\cite{musa2024analyzing,pooch2020can,zech2018variable}. The resulting domain shift can severely degrade diagnostic accuracy, limiting the real-world adoption of automated systems~\cite{ghafoorian2017transfer}. Addressing this performance gap under domain shift remains a key challenge for reliable clinical AI deployment.

To address domain shift, prior work has explored domain adaptation~\cite{ganin2015unsupervised} and domain generalization~\cite{zhou2021mixstyle}, both of which require access to source data or control over the training process. In practice, pretrained models are frequently distributed to new sites where source data is unavailable due to privacy regulations and institutional policies. Test-time adaptation (TTA) provides a more realistic alternative by updating a pretrained model at inference using only unlabeled target data, without any source data or target annotations~\cite{wang2021tent}.

\begin{figure}[!ht]
    \centering
    \includegraphics[width=0.9\linewidth]{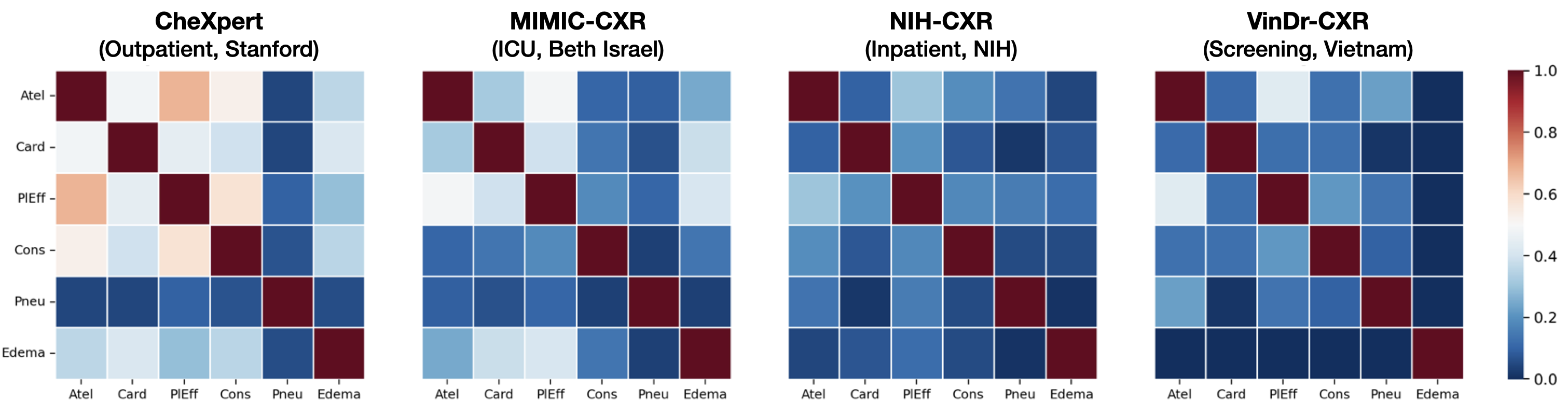}
    \caption{Co-occurrence patterns vary across chest X-ray domains.}
    \label{fig:domain_oracle}
\end{figure}

While TTA has been widely studied for natural image classification, its direct application to multi-label medical imaging remains problematic. Standard TTA minimizes prediction entropy per class and weights all samples equally, implicitly assuming label independence. However, clinical pathologies follow structured co-occurrence patterns rather than occurring in isolation~\cite{wang2017chestx,yao2017learning}. Fig.~\ref{fig:domain_oracle} shows that these dependency structures vary across datasets, reflecting differences in clinical setting. Predictions that violate plausible label relationships may therefore be unreliable despite being individually confident. Yet existing entropy-based TTA methods adapt using all predictions indiscriminately, amplifying inconsistent patterns and introducing noisy gradients under domain shift.

In this work, we propose \textbf{CoWA} (\textbf{Co}-occurrence \textbf{W}eighted \textbf{A}daptation) to bridge this gap. CoWA estimates the pathology co-occurrence matrix from incoming test predictions, capturing the evolving co-occurrence patterns of the target domain without accessing source data. This matrix serves as a reliability signal that measures how well each sample’s predicted label configuration aligns with the emerging target structure. During adaptation, CoWA assigns higher weights to structurally consistent samples and downweights predictions that violate plausible label relationships. By steering updates toward coherent label patterns, CoWA reduces noisy gradients and enables TTA to adapt while respecting structured dependencies in multi-label clinical data, without requiring additional supervision.

Our main contributions can be summarized as follows:
\begin{enumerate}
    \item We propose CoWA, a source-free TTA method that estimates co-occurrence patterns from model predictions and uses them as a per-sample reliability signal for multi-label medical image classification.
    \item We show that standard TTA methods, originally designed for single-label benchmarks, can be ineffective or even harmful when applied to multi-label clinical tasks, highlighting the need for label-structure aware adaptation.
    \item We validate CoWA on multiple domain shift scenarios using public chest X-ray datasets, showing consistent gains over established baselines.
\end{enumerate}

\section{Related Work}
Test-time adaptation (TTA) has emerged as an effective strategy for addressing distribution shifts, as it adapts models to the target distribution without access to source data, unlike traditional domain adaptation.
Existing TTA methods update model parameters via prediction-based objectives. TENT~\cite{wang2021tent} minimizes prediction entropy by updating BN affine parameters, while EATA~\cite{niu2022efficient} extends this by filtering unreliable samples and applying Fisher regularization to mitigate catastrophic forgetting. CoTTA~\cite{wang2022continual} and RoTTA~\cite{yuan2023robust} further address continual adaptation under temporally correlated streams via augmentation-based pseudo-labels and robust BN estimation, respectively. Beyond gradient-based optimization, AdaBN~\cite{li2018adaptive} replaces source BN statistics with those estimated from target samples, serving as a simple yet effective domain correction baseline.
While these methods have shown success on single-label natural image benchmarks such as ImageNet-C~\cite{hendrycks2019benchmarking}, they typically treat all samples equally during adaptation and assume label independence. In multi-label medical imaging, however, chest pathologies exhibit structured co-occurrence patterns~\cite{milne1985radiologic,cardinale2014effectiveness}, suggesting that label dependencies can provide valuable guidance for adaptation.

\section{Proposed Method}

Given a source-pretrained multi-label classifier $f_\theta$ and an unlabeled target dataset $\mathcal{D}_t = \{x_i\}_{i=1}^{N}$, CoWA adapts $f_\theta$ to the target distribution by incorporating target domain label co-occurrence structure into entropy minimization. Instead of uniformly minimizing entropy over all predictions, CoWA estimates the target co-occurrence structure from model outputs and assigns a weight to each sample based on its structural consistency. These weights are then used in entropy minimization during adaptation.

\begin{figure}[!ht]
    \centering
    \includegraphics[width=0.9\linewidth]{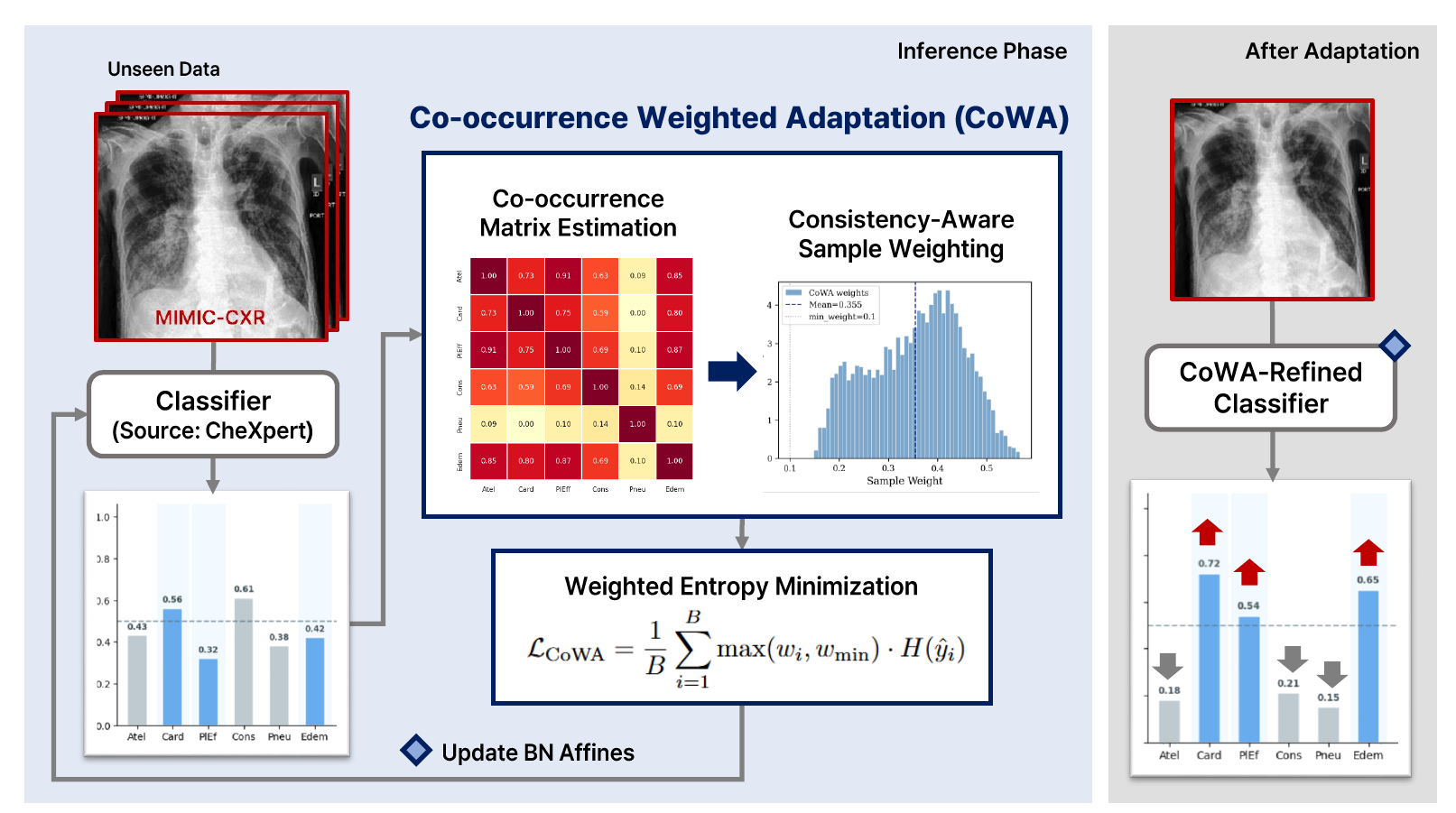}
    \caption{\textbf{Overview of CoWA}. CoWA estimates a pathology co-occurrence matrix from target predictions, computes sample-wise consistency scores, and uses them to weight samples for test-time domain adaptation.}
    \label{fig:framework}
\end{figure}

CoWA consists of three components:
(1) estimation of the target domain co-occurrence matrix from model predictions,
(2) computation of a sample-wise consistency score, and
(3) weighted entropy minimization to update model parameters.
The overall framework is illustrated in Fig.~\ref{fig:framework}.



\paragraph{\textbf{Co-occurrence Matrix Estimation.}}
For an input $x_i$, the model outputs $\hat{y}_i = f_\theta(x_i) \in [0,1]^C$, where $C$ denotes the number of pathologies. We treat these predictions as soft pseudo-labels and obtain $\tilde{y}_i \in \{0,1\}^C$ via fixed threshold binarization to suppress low confidence activations and reduce noise in dependency estimation. Based on these binary predictions, we compute the pairwise co-occurrence matrix:
\begin{equation}
    \mathbf{S} = \sum_{i=1}^{n} \tilde{y}_i \tilde{y}_i^\top,
\end{equation}

\noindent
where $n$ is the number of processed samples and $\mathbf{S}_{jk}$ counts how often pathologies $j$ and $k$ are jointly predicted as positive.
We then compute the empirical joint probabilities $\mathbf{P} = \mathbf{S} / n$ and define the normalized co-occurrence matrix as
\begin{equation}
    {\mathbf{M}}_{jk} = \frac{\mathbf{P}_{jk}}{\sqrt{\mathbf{P}_{jj} \mathbf{P}_{kk}}+\varepsilon}, 
    \quad {\mathbf{M}}_{jj} = 1.
\end{equation}

Here, $\mathbf{S}$ is accumulated across all batches processed so far rather than recomputed per batch, so that $\mathbf{M}$ is progressively refined as adaptation proceeds and the influence of noisy early-stage predictions is dampened.

\paragraph{\textbf{Consistency-aware Sample Weighting.}}
Using the estimated co-occurrence matrix ${\mathbf{M}}$, we assign a consistency score to each target prediction. Here, the consistency of a prediction measures how well its implied label relationships match the estimated target co-occurrence structure. For a sample with predicted probabilities $\hat{y}_i$, we construct a local co-occurrence pattern $\mathbf{m}_i = \hat{y}_i \hat{y}_i^\top$ and compare it with ${\mathbf{M}}$ using the Frobenius norm, converting the deviation into a reliability weight:
\begin{equation}
w_i = \exp\left(-\frac{\|\mathbf{m}_i - {\mathbf{M}}\|_F^2}{\tau}\right),
\label{eq:weight}
\end{equation}
where $\tau$ is a temperature parameter controlling the weighting. Predictions with higher consistency get weights close to 1, while lower-consistency predictions are assigned smaller weights, effectively reducing the influence of samples that are less aligned with the target co-occurrence structure.

\paragraph{\textbf{Weighted Entropy Minimization.}} 
Existing entropy-based TTA methods~\cite{wang2021tent,niu2022efficient} adapt models by minimizing the uniformly averaged prediction entropy, treating all test samples equally. This implicitly assumes that every confident prediction is equally reliable, regardless of its structural consistency with the target domain. In contrast, CoWA reweights each sample based on its consistency score and adjusts its contribution to the adaptation objective:

\begin{equation} 
\mathcal{L}_\text{CoWA} = \frac{1}{B} \sum_{i=1}^{B} \max(w_i, w_{\min}) \cdot H(\hat{y}_i), 
\end{equation} 
where $H(\cdot)$ denotes the entropy of the model prediction, $B$ is the batch size, and $w_{\min}$ prevents the weights from collapsing to zero for numerical stability, acting as an early-stage safeguard before $\mathbf{M}$ stabilizes. Samples whose predictions align well with the estimated target co-occurrence structure receive larger weights and thus exert stronger influence on adaptation, whereas structurally inconsistent samples are suppressed. Following prior work~\cite{wang2021tent}, we adapt only the BN affine parameters of $f_\theta$.

\section{Experiments}

\subsection{Experimental Setup}
\textbf{Datasets.}
We evaluate on four public chest X-ray datasets: MIMIC-CXR~\cite{johnson2019mimic}, CheXpert~\cite{irvin2019chexpert}, VinDr-CXR~\cite{nguyen2022vindr}, and NIH Chest X-ray~\cite{wang2017chestx}. We construct six source--target pairs to cover diverse domain shift scenarios. For each pair, the unseen test split of the target domain is used for evaluation. Results are reported on six shared pathologies: Atelectasis (At.), Cardiomegaly (Ca.), Pleural Effusion (Ef.), Consolidation (Co.), Pneumothorax (Pt.), and Edema (Ed.).

\noindent\textbf{Pretrained Classifiers.} 
We use DenseNet-121 backbones from TorchXRayVision~\cite{cohen2022torchxrayvision}, pretrained on NIH and CheXpert. Each pretrained model is treated as a distinct source model when evaluating cross-domain shifts.

\noindent\textbf{Baselines.} 
We evaluate CoWA against \textit{No Adapt.} (source model without adaptation), AdaBN~\cite{li2018adaptive}, and four TTA methods: TENT~\cite{wang2021tent}, CoTTA~\cite{wang2022continual}, EATA~\cite{niu2022efficient}, and RoTTA~\cite{yuan2023robust}.

\noindent\textbf{Implementation Details.} 
All methods use a batch size of 64, with baseline hyperparameters following their original papers. For CoWA, the learning rate is selected from [1e-4, 1e-2], with $\tau \in \{0.01, 0.05, 0.1, 0.5\}$, and the co-occurrence threshold $\in \{0.4, 0.5, 0.6\}$. We fix $w_{\min}=0.01$ in all settings. All experiments are conducted on four NVIDIA GeForce RTX 3090 GPUs.


\subsection{Main Results}

Table~\ref{tab:da_results} reports the AUROC across six domain shift scenarios. CoWA achieves the best mean performance in most settings and remains competitive in the others, consistently ranking first or second with only small gaps. In contrast, TENT and AdaBN show clear variability across domain pairs, performing well in some cases but degrading substantially in others. CoTTA and RoTTA often remain close to the unadapted baseline under larger shifts. Class-wise results further explain these trends. Competing methods frequently suffer marked drops on low-prevalence pathologies (e.g., Pneumothorax), where entropy minimization or simple BN updates can reinforce unreliable predictions. CoWA consistently alleviates such class-specific degradations, indicating that its higher mean AUROC is driven by preventing severe failures on vulnerable classes. 
In clinical settings, where a severe drop on any single pathology can compromise diagnosis, such worst-case robustness is as important as mean performance: notably, CoWA is the only method that avoids severe drops below the unadapted baseline across all shifts.

\begin{table*}[!t]
\renewcommand{\arraystretch}{1.2} 
\centering
\fontsize{8}{10}\selectfont
\definecolor{sourcebg}{RGB}{210, 230, 255}   
\definecolor{cowabg}{RGB}{255,245,210}   
\caption{AUROC under multiple domain-shift settings across source domains. \textbf{Bold} indicates the best result and \underline{underline} the second best. Values are scaled by 100 for readability.}
\label{tab:da_results}
\begin{tabular}{l| ccccccc|| ccccccc}
\hline

\multirow{3}{*}{\textbf{Method}} 
& \multicolumn{7}{c||}{\textbf{Source: CheXpert}} 
& \multicolumn{7}{c}{\textbf{Source: NIH}} \\
\cline{2-8} \cline{9-15}
& \multicolumn{6}{c}{\textbf{Pathology}} & \multirow{2}{*}{\textbf{Mean}} 
& \multicolumn{6}{c}{\textbf{Pathology}} & \multirow{2}{*}{\textbf{Mean}} \\
\cline{2-7} \cline{9-14}
& At. & Ca. & Ef. & Co. & Pt. & \multicolumn{1}{c}{Ed.} &  
& At. & Ca. & Ef. & Co. & Pt. & \multicolumn{1}{c}{Ed.} &   \\
\hline
\rowcolor{gray!20} &\multicolumn{14}{c}{\rule[-0.5ex]{0pt}{3ex}\textbf{Target: MIMIC}} \\
\hline
No Adapt. & 61.8 & \underline{69.5} & 73.8 & 58.7 & 63.0 & 73.6 & \cellcolor{cowabg}66.7 & 62.6 & 66.3 & \underline{73.9} & 57.0 & \textbf{57.5} & 67.8 & \cellcolor{cowabg}64.2 \\
AdaBN     & 63.4 & 68.5 & \textbf{76.0} & \textbf{62.2} & \underline{65.9} & 72.8 & \cellcolor{cowabg}\underline{68.1} & 63.5 & \underline{68.9} & \underline{73.9} & \underline{57.1} & 48.3 & 69.7 & \cellcolor{cowabg}63.6 \\
TENT      & \textbf{64.8} & 68.0 & 75.5 & 59.4 & 64.5 & \underline{73.7} & \cellcolor{cowabg}67.6 & 62.8 & 65.2 & 71.2 & 54.8 & 49.1 & 62.8 & \cellcolor{cowabg}61.0 \\
CoTTA     & 61.9 & \underline{69.5} & 73.8 & 58.6 & 63.0 & 73.6 & \cellcolor{cowabg}66.7 & 62.6 & 66.4 & \underline{73.9} & 57.0 & \underline{57.3} & 67.9 & \cellcolor{cowabg}64.2 \\
EATA      & 63.3 & 68.5 & \underline{75.6} & \underline{61.5} & 65.8 & 72.6 & \cellcolor{cowabg}67.9 & \underline{64.6} & \textbf{69.6} & \textbf{74.8} & \textbf{57.8} & 49.5 & \underline{72.0} & \cellcolor{cowabg}\underline{64.7} \\
RoTTA     & 61.8 & \textbf{69.6} & 73.8 & 58.6 & 62.9 & 73.6 & \cellcolor{cowabg}66.7 & 62.5 & 66.2 & 73.8 & 56.9 & \textbf{57.5} & 67.7 & \cellcolor{cowabg}64.1 \\ 
\hline
\textbf{CoWA} &  \underline{64.2} & 68.6 & \textbf{76.0} & 61.3 & \textbf{66.3} & \textbf{73.8} & \cellcolor{cowabg} \textbf{68.4} & \textbf{65.1} & \underline{68.9} & 73.3 & \textbf{57.8} & 55.7 & \textbf{72.1} & \cellcolor{cowabg}\textbf{65.5} \\

\hline
\rowcolor{gray!20} & \multicolumn{14}{c}{\rule[-0.5ex]{0pt}{3ex}\textbf{Target: VinDr}} \\
\hline
No Adapt. & 77.7 & \textbf{85.3} & 82.6 & 81.0 & 75.7 & -- & \cellcolor{cowabg}80.5 & 74.9 & 82.4 & 85.0 & 76.5 & 84.4 & -- & \cellcolor{cowabg}80.6 \\
AdaBN     & \underline{80.6} & \underline{84.6} & \underline{87.1} & \underline{82.8} & \underline{87.7} & -- & \cellcolor{cowabg}\underline{84.6} & {79.8} & \textbf{87.9} & 86.1 & {78.0} & 86.4 & -- & \cellcolor{cowabg}83.6 \\
TENT      & 70.8 & 75.7 & 77.6 & 75.3 & 72.5 & -- & \cellcolor{cowabg}74.4 & \textbf{83.6} & \underline{86.5} & \underline{88.3} & \textbf{78.5} & \textbf{87.4} & -- & \cellcolor{cowabg}\textbf{84.9} \\ 
CoTTA     & 77.9 & \textbf{85.3} & 82.8 & 81.2 & 76.6 & -- & \cellcolor{cowabg}80.8 & 75.0 & 82.5 & 85.1 & 76.6 & 84.5 & -- & \cellcolor{cowabg}80.7 \\
EATA      & 79.2 & 83.5 & 86.5 & {82.5} & 87.6 & -- & \cellcolor{cowabg}83.9 & 68.7 & 80.2 & 80.5 & 72.7 & 74.5 & -- & \cellcolor{cowabg}75.3 \\
RoTTA     & 77.4 & 85.1 & 82.3 & 80.9 & 74.9 & -- & \cellcolor{cowabg}80.1 & 74.9 & 82.4 & 85.0 & 76.4 & 84.2 & -- & \cellcolor{cowabg}80.6 \\
\hline
\textbf{CoWA} & \textbf{82.0} & \underline{84.6} & \textbf{88.1} & \textbf{83.6} & \textbf{89.9} & -- & \cellcolor{cowabg}\textbf{85.6} & \underline{83.4} & 83.6 & \textbf{88.8} & \underline{78.2} & \underline{87.1} & -- & \cellcolor{cowabg}\underline{84.2} \\

\hline
\rowcolor{gray!20} & \multicolumn{7}{c||}{\rule[-0.5ex]{0pt}{3ex}\textbf{Target: NIH}} & \multicolumn{7}{c}{\textbf{Target: CheXpert}} \\
\hline
No Adapt. & 56.6 & 62.3 & 67.4 & \underline{67.4} & 64.1 & 69.7 & \cellcolor{cowabg}64.6 & 74.3 & 71.7 & 79.4 & 80.8 & \textbf{71.2} & 78.3 & \cellcolor{cowabg}76.0 \\
AdaBN     & \underline{59.4} & 63.9 & \textbf{71.5} & \textbf{67.7} & \textbf{69.4} & 70.0 & \cellcolor{cowabg}\textbf{67.0} & \textbf{80.2} & \textbf{79.9} & 78.5 & 81.9 & 56.7 & 79.9 & \cellcolor{cowabg}76.2 \\
TENT      & 50.1 & 50.1 & 53.5 & 55.3 & 52.0 & 63.0 & \cellcolor{cowabg}54.0 & 77.7 & 76.5 & \underline{80.6} & \underline{83.2} & 65.5 & \underline{81.1} & \cellcolor{cowabg}\underline{77.4} \\
CoTTA     & 56.3 & 55.4 & 59.3 & 61.2 & \underline{66.6} & 64.8 & \cellcolor{cowabg}60.6 & 74.3 & 71.7 & 79.4 & 80.8 & \underline{71.1} & 78.3 & \cellcolor{cowabg}75.9 \\
EATA      & 59.3 & \textbf{67.0} & 70.5 & {67.0} & 63.7 & \underline{70.1} & \cellcolor{cowabg}66.3 & 77.9 & 77.8 & \textbf{80.9} & 82.9 & 60.6 & \textbf{81.3} & \cellcolor{cowabg}76.9 \\
RoTTA     & 56.8 & 63.3 & 66.2 & 66.7 & 59.5 & \underline{70.1} & \cellcolor{cowabg}63.8 & 74.3 & 71.7 & 79.4 & 80.8 & \textbf{71.2} & 78.3 & \cellcolor{cowabg}75.9 \\
\hline
\textbf{CoWA} & \textbf{59.7} & \underline{66.8} & \underline{71.1} & {67.3} & 65.4 & \textbf{70.7} & \cellcolor{cowabg}\underline{66.8} & \underline{79.2} & \underline{78.1} & 80.5 & \textbf{83.7} & 66.1 & 79.6 & \cellcolor{cowabg}\textbf{77.9} \\

\hline
\end{tabular}
\end{table*}

\subsection{Analysis}
We analyze CoWA from three perspectives: (1) the quality of the estimated co-occurrence matrix, (2) the reliability of the resulting sample weights, and (3) an ablation study on the norm choice in the weighting mechanism.

\begin{figure}[!ht]
    \centering
    \includegraphics[width=0.9\linewidth]{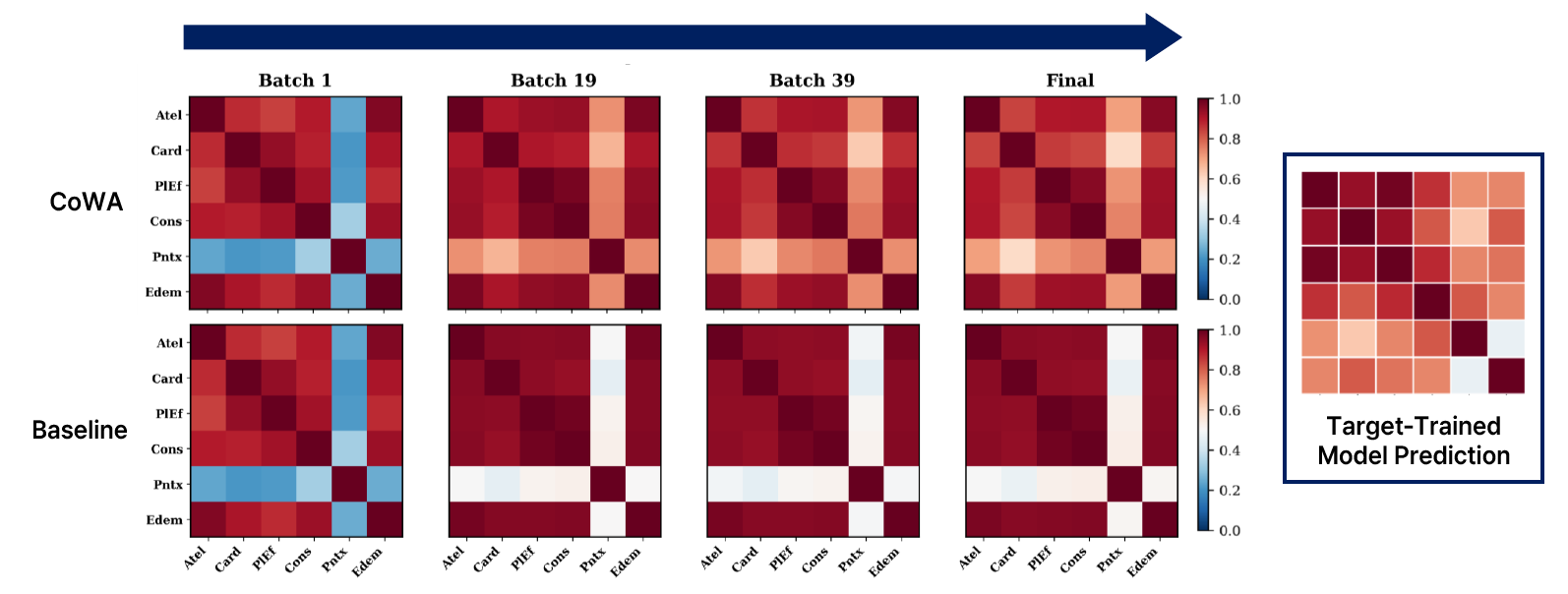}
    \caption{\textbf{Co-occurrence matrix evolution} (CheXpert $\to$ MIMIC). CoWA (top) gradually aligns toward the target-trained reference (right), while the baseline (bottom) progressively loses clear dependency structure over batches.}
    \label{fig:matrix_evolution}
\end{figure}

\paragraph{\textbf{Does the Estimated Matrix Preserve Target Label Structure?}} 
CoWA assumes that the estimated co-occurrence matrix ${\mathbf{M}}$ captures the label dependency structure of the target domain during adaptation. To evaluate this assumption, we construct a reference matrix by passing target-domain test data through the same architecture trained with full supervision on the target domain, which serves as a proxy for the true target co-occurrence structure. Fig.~\ref{fig:matrix_evolution} illustrates how ${\mathbf{M}}$ evolves under the CheXpert $\to$ MIMIC shift. At Batch 1, CoWA and the baseline are identical, as both models generate predictions from the source-trained model without adaptation. As adaptation proceeds, however, their behaviors diverge. With CoWA (top), the estimated matrix progressively aligns with the reference matrix. In contrast, the baseline, which does not update during testing, increasingly deviates from the reference in co-occurrence patterns involving Pneumothorax. In particular, its co-occurrence values between Pneumothorax and other pathologies become less distinguishable, resulting in blurred dependency structure. These results indicate that co-occurrence-guided weighting better preserves target-domain label dependencies during adaptation than the baseline.

\paragraph{\textbf{Do CoWA Weights Reflect Sample Reliability?}} 

\begin{figure}[!ht]
    \centering
    \includegraphics[width=0.6\linewidth]{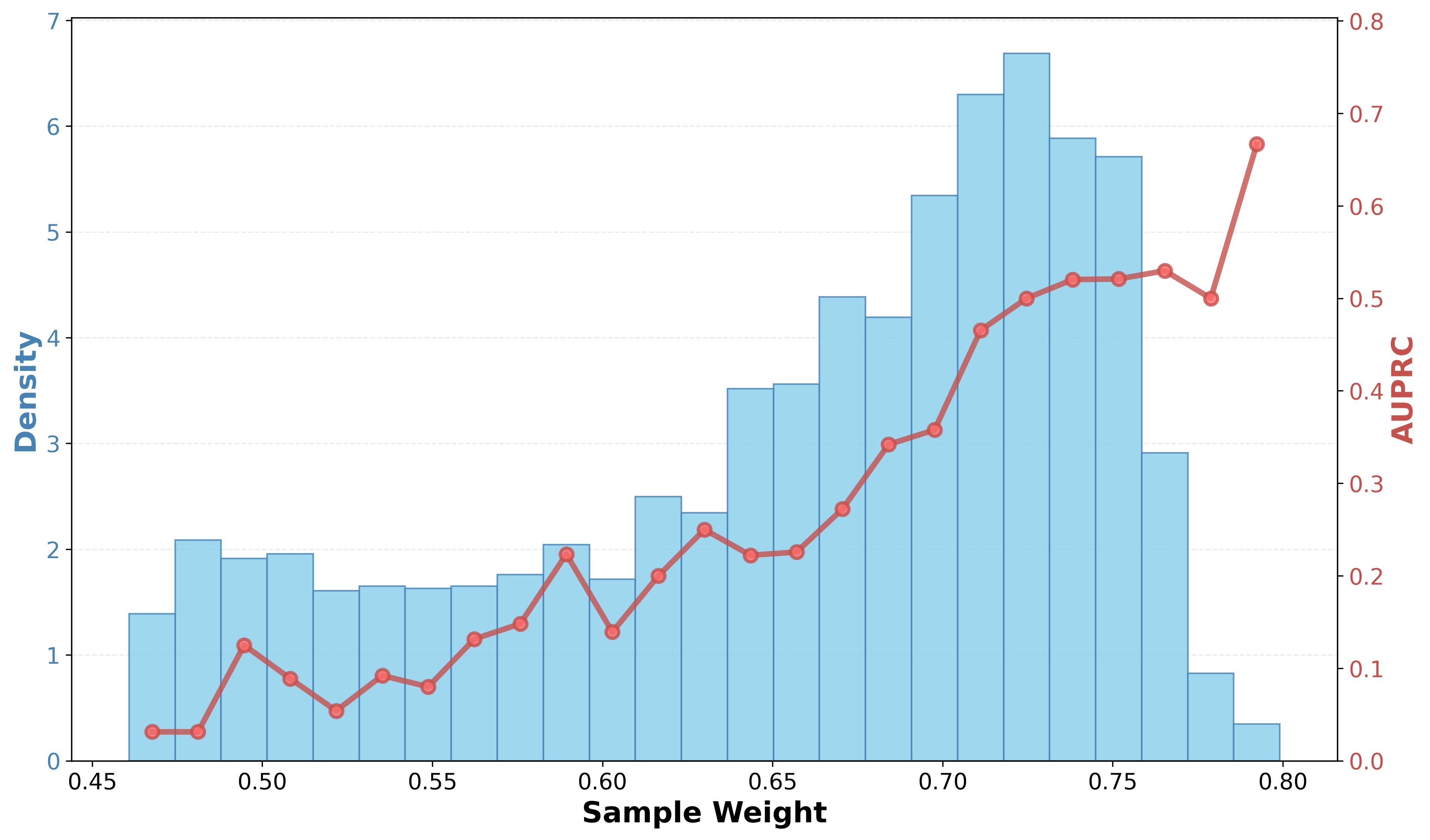}
    \caption{\textbf{Sample weight distribution and prediction reliability.} Blue bars show the weight distribution after adaptation, and red line shows AUPRC per weight bin. Higher-weight samples consistently achieve better prediction quality.}
    \label{fig:weight_dist}
\end{figure}

To examine whether CoWA weights reflect sample reliability, we analyze prediction quality as a function of weight under the CheXpert $\rightarrow$ MIMIC domain shift. Target samples are grouped into weight bins, and classification performance within each bin is computed using ground-truth labels. We report AUPRC due to the severe class imbalance in chest X-ray datasets~\cite{saito2015precision}. Fig.~\ref{fig:weight_dist} presents the weight distribution from the final adaptation batch (blue) together with the corresponding performance for each bin (red). A positive association is observed between weight magnitude and predictive performance: lower-weight samples exhibit weaker predictive quality, whereas higher-weight samples achieve stronger performance. These findings suggest that co-occurrence consistency correlates with sample reliability, supporting the role of CoWA weights as an implicit reliability estimator during adaptation.

\paragraph{\textbf{Ablation Study}.} 
We first verify that the proposed reweighting stabilizes adaptation: the variance of the BN affine gradient $\ell_2$-norm under CoWA is about half that of TENT ($1.02\times10^{-5}$ vs.\ $2.38\times10^{-5}$), confirming that consistency-based reweighting yields more stable updates.
We then justify the choice of the Frobenius norm ($\ell_F$) in Eq.~\ref{eq:weight} by comparing it against $\ell_1$, $\ell_2$, and $\ell_\infty$ alternatives over all six domain shift scenarios (Table~\ref{tab:ablation_norm}). $\ell_F$ achieves the highest mean AUROC while also exhibiting the lowest variance across the six shifts, indicating more stable adaptation. Unlike $\ell_1$, $\ell_2$, or $\ell_\infty$, which can be dominated by large deviations in a few individual entries, $\ell_F$ aggregates squared differences over all entries and thus better captures the overall discrepancy of the co-occurrence matrix, making it more appropriate for aligning label dependencies rather than individual entries.


\begin{table}[!ht]
\centering
\setlength{\arrayrulewidth}{0.4pt}
\setlength{\tabcolsep}{4pt} 
\caption{AUROC of norm ablation in CoWA across six domain-shift scenarios.}
\label{tab:ablation_norm}
\resizebox{0.9\columnwidth}{!}{%
\begin{tabular}{c|ccc|ccc|c}
\hline
\multirow{2}{*}{\textbf{Norm}} 
& \multicolumn{3}{c|}{\textbf{Source: NIH}} 
& \multicolumn{3}{c|}{\textbf{Source: CheXpert}} 
& \multirow{2}{*}{\textbf{Mean}} \\
\cline{2-7}
& \textbf{CheXpert} & \textbf{MIMIC} & \textbf{VinDr} 
& \textbf{NIH} & \textbf{MIMIC} & \textbf{VinDr} & \\
\hline
$\ell_1$          & 77.9 & 51.8 & 84.2 & 59.5 & 68.4 & 85.0 & 71.1 ($\pm$12.2) \\
\hline
$\ell_2$          & 77.8 & 50.6 & 84.2 & 59.4 & 67.0 & 79.7 & 69.8 ($\pm$11.8) \\
\hline
$\ell_\infty$     & 77.9 & 50.6 & 84.2 & 59.4 & 68.4 & 83.2 & 70.6 ($\pm$12.3) \\
\hline
\textbf{$\ell_F$} & \textbf{77.9} & \textbf{65.5} & \textbf{84.2} & \textbf{66.8} & \textbf{68.4} & \textbf{85.6} & \textbf{74.7 ($\pm$8.0)} \\
\hline
\end{tabular}%
}
\end{table}

\section{Conclusion}

We propose CoWA, a TTA method that leverages disease co-occurrence patterns estimated from model predictions as a per-sample reliability signal for entropy minimization. Unlike standard TTA approaches that treat all samples equally without considering label dependencies, CoWA downweights predictions with inconsistent label combinations, enabling adaptation to focus on structurally consistent samples. Experiments across multiple chest X-ray domain shift scenarios demonstrate consistent improvements over established baselines. By incorporating target-domain co-occurrence structure into TTA without requiring additional annotations or architectural changes, CoWA offers a simple and deployable solution for improving diagnostic robustness under domain shift. As the estimated patterns may be sensitive to dataset-level biases, extending CoWA to capture more robust, causally grounded relationships is a promising future direction.

\begin{credits}
\subsubsection{\ackname}
This work was supported in part by the Institute of Information \& Communications Technology Planning \& Evaluation (IITP) (Grant No. RS-2022-II220984) and in part by the National Research Foundation of Korea (NRF) (Grant No. RS-2024-00338859, RS-2025-00515481), both funded by the Ministry of Science and ICT (MSIT), Republic of Korea.

\subsubsection{\discintname}
The authors have no competing interests to declare that are relevant to the content of this article.
\end{credits}

\bibliographystyle{splncs04}
\bibliography{Paper-4786}

@inproceedings{wang2017chestx,
  title={Chestx-ray8: Hospital-scale chest x-ray database and benchmarks on weakly-supervised classification and localization of common thorax diseases},
  author={Wang, Xiaosong and Peng, Yifan and Lu, Le and Lu, Zhiyong and Bagheri, Mohammadhadi and Summers, Ronald M},
  booktitle={Proceedings of the IEEE conference on computer vision and pattern recognition},
  pages={2097--2106},
  year={2017}
}

@article{johnson2019mimic,
  title={Mimic-cxr-jpg-chest radiographs with structured labels},
  author={Johnson, Alistair and Lungren, Matt and Peng, Yifan and Lu, Zhiyong and Mark, Roger and Berkowitz, Seth and Horng, Steven},
  journal={PhysioNet},
  volume={101},
  pages={215--220},
  year={2019}
}

@article{nguyen2022vindr,
  title={VinDr-CXR: An open dataset of chest X-rays with radiologist’s annotations},
  author={Nguyen, Ha Q and Lam, Khanh and Le, Linh T and Pham, Hieu H and Tran, Dat Q and Nguyen, Dung B and Le, Dung D and Pham, Chi M and Tong, Hang TT and Dinh, Diep H and others},
  journal={Scientific Data},
  volume={9},
  number={1},
  pages={429},
  year={2022},
  publisher={Nature Publishing Group UK London}
}

@inproceedings{irvin2019chexpert,
  title={Chexpert: A large chest radiograph dataset with uncertainty labels and expert comparison},
  author={Irvin, Jeremy and Rajpurkar, Pranav and Ko, Michael and Yu, Yifan and Ciurea-Ilcus, Silviana and Chute, Chris and Marklund, Henrik and Haghgoo, Behzad and Ball, Robyn and Shpanskaya, Katie and others},
  booktitle={Proceedings of the AAAI conference on artificial intelligence},
  volume={33},
  pages={590--597},
  year={2019}
}

@inproceedings{wang2021tent,
  title={Tent: Fully Test-Time Adaptation by Entropy Minimization},
  author={Wang, Dequan and Shelhamer, Evan and Liu, Shaoteng and Olshausen, Bruno and Darrell, Trevor},
  booktitle={International Conference on Learning Representations},
  year={2021}
}

@inproceedings{wang2022continual,
  title={Continual Test-Time Domain Adaptation},
  author={Wang, Qin and Fink, Olga and Van Gool, Luc and Dai, Dengxin},
  booktitle={Proceedings of the IEEE/CVF Conference on Computer Vision and Pattern Recognition},
  pages={7201--7211},
  year={2022}
}

@article{li2018adaptive,
  title={Adaptive batch normalization for practical domain adaptation},
  author={Li, Yanghao and Wang, Naiyan and Shi, Jianping and Hou, Xiaodi and Liu, Jiaying},
  journal={Pattern Recognition},
  volume={80},
  pages={109--117},
  year={2018},
  publisher={Elsevier}
}

@inproceedings{niu2022efficient,
  title={Efficient Test-Time Model Adaptation without Forgetting},
  author={Niu, Shuaicheng and Wu, Jiaxiang and Zhang, Yifan and Chen, Yaofo and Zheng, Shijian and Zhao, Peilin and Tan, Mingkui},
  booktitle={Proceedings of the 39th International Conference on Machine Learning},
  pages={16888--16905},
  year={2022},
  publisher={PMLR}
}

@inproceedings{yuan2023robust,
  title={Robust Test-Time Adaptation in Dynamic Scenarios},
  author={Yuan, Longhui and Xie, Binhui and Li, Shuang},
  booktitle={Proceedings of the IEEE/CVF Conference on Computer Vision and Pattern Recognition},
  pages={15922--15932},
  year={2023}
}

@InProceedings{cohen2022torchxrayvision,
  title={TorchXRayVision: A library of chest X-ray datasets and models},
  author={Cohen, Joseph Paul and Viviano, Joseph D. and Bertin, Paul and Morrison, Paul and Torabian, Parsa and Guarrera, Matteo and Lungren, Matthew P and Chaudhari, Akshay and Brooks, Rupert and Hashir, Mohammad and Bertrand, Hadrien},
  booktitle={Proceedings of The 5th International Conference on Medical Imaging with Deep Learning},
  pages={231--249},
  year={2022},
  volume={172},
  series={Proceedings of Machine Learning Research},
  publisher={PMLR}
}

@article{zech2018variable,
  title={Variable generalization performance of a deep learning model to detect pneumonia in chest radiographs: A cross-sectional study},
  author={Zech, John R and Badgeley, Marcus A and Liu, Manway and Costa, Anthony B and Titano, Joseph J and Oermann, Eric Karl},
  journal={PLoS Medicine},
  volume={15},
  number={11},
  pages={e1002683},
  year={2018},
  publisher={Public Library of Science}
}

@inproceedings{pooch2020can,
  title={Can We Trust Deep Learning Based Diagnosis? The Impact of Domain Shift in Chest Radiograph Classification},
  author={Pooch, Eduardo H. P. and Ballester, Pedro L. and Barros, Rodrigo C.},
  booktitle={Thoracic Image Analysis (TIA 2020), Lecture Notes in Computer Science},
  volume={12502},
  pages={74--83},
  year={2020},
  publisher={Springer}
}

@inproceedings{musa2024analyzing,
  title={Analyzing Cross-Population Domain Shift in Chest X-Ray Image Classification and Mitigating the Gap with Deep Supervised Domain Adaptation},
  author={Musa, Aminu and Ibrahim Adamu, Mubarak and Kakudi, Habeebah Adamu and Hernandez, Monica and Lawal, Yusuf},
  booktitle={International Conference on Medical Image Computing and Computer-Assisted Intervention},
  pages={585--595},
  year={2024},
  publisher={Springer}
}

@inproceedings{ganin2015unsupervised,
  title={Unsupervised Domain Adaptation by Backpropagation},
  author={Ganin, Yaroslav and Lempitsky, Victor},
  booktitle={International Conference on Machine Learning},
  pages={1180--1189},
  year={2015},
  publisher={PMLR}
}

@inproceedings{zhou2021mixstyle,
  title={Domain Generalization with MixStyle},
  author={Zhou, Kaiyang and Yang, Yongxin and Qiao, Yu and Xiang, Tao},
  booktitle={International Conference on Learning Representations},
  year={2021}
}

@inproceedings{hendrycks2019benchmarking,
  title={Benchmarking Neural Network Robustness to Common Corruptions and Perturbations},
  author={Hendrycks, Dan and Dietterich, Thomas},
  booktitle={International Conference on Learning Representations},
  year={2019}
}

@article{milne1985radiologic,
  title={The radiologic distinction of cardiogenic and noncardiogenic edema},
  author={Milne, EN and Pistolesi, Massimo and Miniati, Massimo and Giuntini, Carlo},
  journal={American journal of roentgenology},
  volume={144},
  number={5},
  pages={879--894},
  year={1985},
  publisher={American Roentgen Ray Society}
}

@article{cardinale2014effectiveness,
  title={Effectiveness of chest radiography, lung ultrasound and thoracic computed tomography in the diagnosis of congestive heart failure},
  author={Cardinale, Luciano and Priola, Adriano Massimiliano and Moretti, Federica and Volpicelli, Giovanni},
  journal={World journal of radiology},
  volume={6},
  number={6},
  pages={230},
  year={2014}
}

@inproceedings{ghafoorian2017transfer,
  title={Transfer learning for domain adaptation in MRI: Application in brain lesion segmentation},
  author={Ghafoorian, Mohsen and Mehrtash, Alireza and Kapur, Tina and Karssemeijer, Nico and Marchiori, Elena and Pesteie, Mehran and Guttmann, Charles RG and De Leeuw, Frank-Erik and Tempany, Clare M and Van Ginneken, Bram and others},
  booktitle={International conference on medical image computing and computer-assisted intervention},
  pages={516--524},
  year={2017},
  organization={Springer}
}

@article{yao2017learning,
  title={Learning to diagnose from scratch by exploiting dependencies among labels},
  author={Yao, Li and Poblenz, Eric and Dagunts, Dmitry and Covington, Ben and Bernard, Devon and Lyman, Kevin},
  journal={arXiv preprint arXiv:1710.10501},
  year={2017}
}

@article{saito2015precision,
  title={The precision-recall plot is more informative than the ROC plot when evaluating binary classifiers on imbalanced datasets},
  author={Saito, Takaya and Rehmsmeier, Marc},
  journal={PloS one},
  volume={10},
  number={3},
  pages={e0118432},
  year={2015},
  publisher={Public Library of Science}
}
%





\end{document}